\documentclass[sigconf]{acmart}
\usepackage{booktabs}
\usepackage{multirow}
\usepackage{subfigure}
\usepackage[utf8]{inputenc}
\usepackage{bm}
\usepackage{diagbox}

\usepackage{xspace}
\usepackage{url}
\usepackage{array}
\usepackage{verbatim}
\usepackage{soul, color}
\usepackage{enumitem}
\usepackage{multirow}
\usepackage{amsmath}
\usepackage{algorithm}
\usepackage{algpseudocode}

\usepackage{balance}
\AtBeginDocument{%
  \providecommand\BibTeX{{%
    \normalfont B\kern-0.5em{\scshape i\kern-0.25em b}\kern-0.8em\TeX}}}
    
\newcommand{\eat}[1]{}
\newcommand{\paratitle}[1]{\vspace{1.5ex}\noindent \textbf{#1}}
\newcommand{\ie}{\emph{i.e.,}\xspace}
\newcommand{\eg}{\emph{e.g.,}\xspace}
\newcommand{\etal}{\emph{et al.}\xspace}

\newcommand{\baby}{\textsc{AESM$^2$}\xspace}

\copyrightyear{2022}
\acmYear{2022}
\setcopyright{acmcopyright}\acmConference[SIGIR '22]{Proceedings of the 45th
International ACM SIGIR Conference on Research and Development in Information
Retrieval}{July 11--15, 2022}{Madrid, Spain}
\acmBooktitle{Proceedings of the 45th International ACM SIGIR Conference on
Research and Development in Information Retrieval (SIGIR '22), July 11--15, 2022,
Madrid, Spain}
\acmPrice{15.00}
\acmDOI{10.1145/3477495.3531942}
\acmISBN{978-1-4503-8732-3/22/07}
\begin{document}
\fancyhead{}

\title{Automatic Expert Selection for Multi-Scenario and Multi-Task Search}\thanks{$^\dagger$Chenliang Li is the corresponding author. Work done when Xinyu Zou was an intern at Ant Group.}
\author{Xinyu Zou}
\email{zouxinyu@whu.edu.cn}
\affiliation{
  \institution{School of Cyber Science and Engineering, Wuhan University}
  \country{China}
}

\author{Zhi Hu}
\email{tuopu.hz@antgroup.com}
\affiliation{
  \institution{Ant Group}
  \country{China}
}

\author{Yiming Zhao}
\email{desert.zym@antgroup.com}
\affiliation{
  \institution{Ant Group}
  \country{China}
}

\author{Xuchu Ding}
\email{dingxuchu.dxc@antgroup.com}
\affiliation{
  \institution{Ant Group}
  \country{China}
}

\author{Zhongyi Liu}
\email{zhongyi.lzy@antgroup.com}
\affiliation{
  \institution{Ant Group}
  \country{China}
}

\author{Chenliang Li$^{\dagger}$}
\email{cllee@whu.edu.cn}
\affiliation{
  \institution{Key Laboratory of Aerospace Information Security and Trusted Computing, Ministry of Education, School of Cyber Science and Engineering, Wuhan University}
  \country{China}
}

\author{Aixin Sun}
\email{axsun@ntu.edu.sg}
\affiliation{
  \institution{Nanyang Technological University}
  \country{Singapore}
}

\def\authors{Xinyu Zou, Zhi Hu, Yiming Zhao, Xuchu Ding, Zhongyi Liu, Chenliang Li, Aixin Sun}
\renewcommand{\shortauthors}{Xinyu Zou, et al.}

\begin{abstract}

Multi-scenario learning (MSL) enables a service provider to cater for users' fine-grained demands by separating services for different user sectors, \eg by user's geographical region. Under each scenario there is a need to optimize multiple task-specific targets \eg click through rate and conversion rate, known as multi-task learning (MTL). Recent solutions for MSL and MTL are mostly based on the multi-gate mixture-of-experts (MMoE) architecture. MMoE structure is typically static and its design requires domain-specific knowledge, making it less effective in handling both MSL and MTL. In this paper, we propose a novel \textbf{A}utomatic \textbf{E}xpert \textbf{S}election framework for \textbf{M}ulti-scenario and \textbf{M}ulti-task search, named \baby. \baby integrates both MSL and MTL into a unified framework with an automatic structure learning. Specifically, \baby stacks multi-task layers over multi-scenario layers. This hierarchical design enables us to flexibly establish intrinsic connections between different scenarios, and at the same time also supports high-level feature extraction for different tasks. At each multi-scenario/multi-task layer, a novel expert selection algorithm is proposed to automatically identify scenario-/task-specific and shared experts for each input. Experiments over two real-world large-scale datasets demonstrate the effectiveness of  \baby over a battery of strong baselines. Online A/B test also shows substantial performance gain on multiple metrics. Currently, \baby has been deployed online for serving major traffic.
\end{abstract}




\maketitle
%====================================
\section{Introduction}
\label{sec:intro}
%====================================
\begin{figure}[!]  
\centering  
\includegraphics[width=0.42\textwidth]{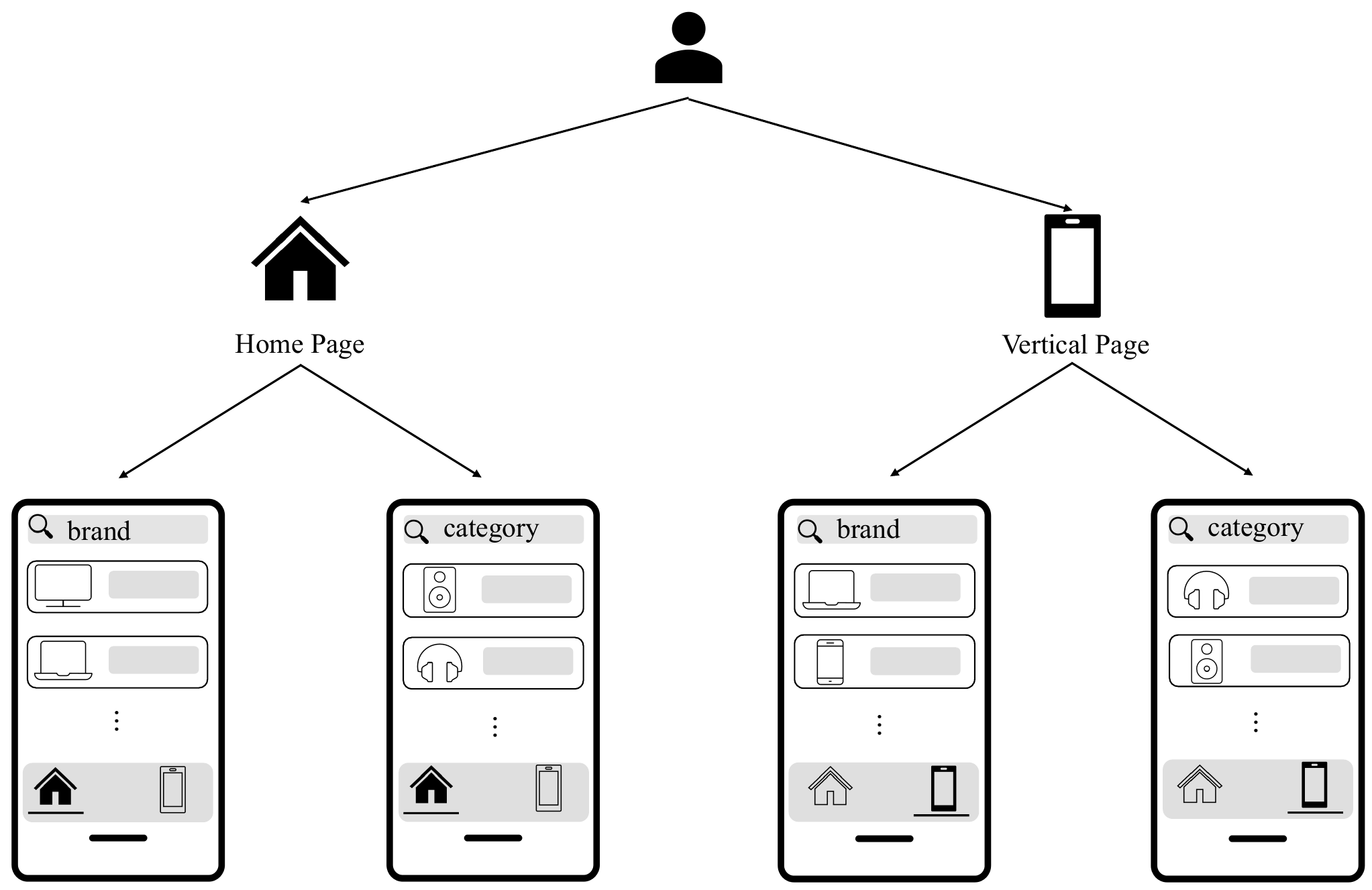}  
\caption{An example of a hierarchical structure of multiple fine-grained scenarios.}
\label{fig:intro}  
\end{figure}

To better serve users, many industrial searching systems start to split business into different scenarios by user sectors, item categories and traffic channels. For example, users may come to a search service through both PC browser and a mobile app, or through a home page or the specific vertical page. The things would become more complex such that fine-grained scenarios are organized as a hierarchical structure. Figure~\ref{fig:intro} shows such a prevalent situation that either brand and category based search can be accessed through home page and vertical page respectively. Although the same search service can be deployed at the back end, the combination of vertical page and brand based search certainly indicates a more focused demand while home page and category based search is more useful for general exploration.

\begin{figure}  
\centering  
\includegraphics[width=.46\textwidth]{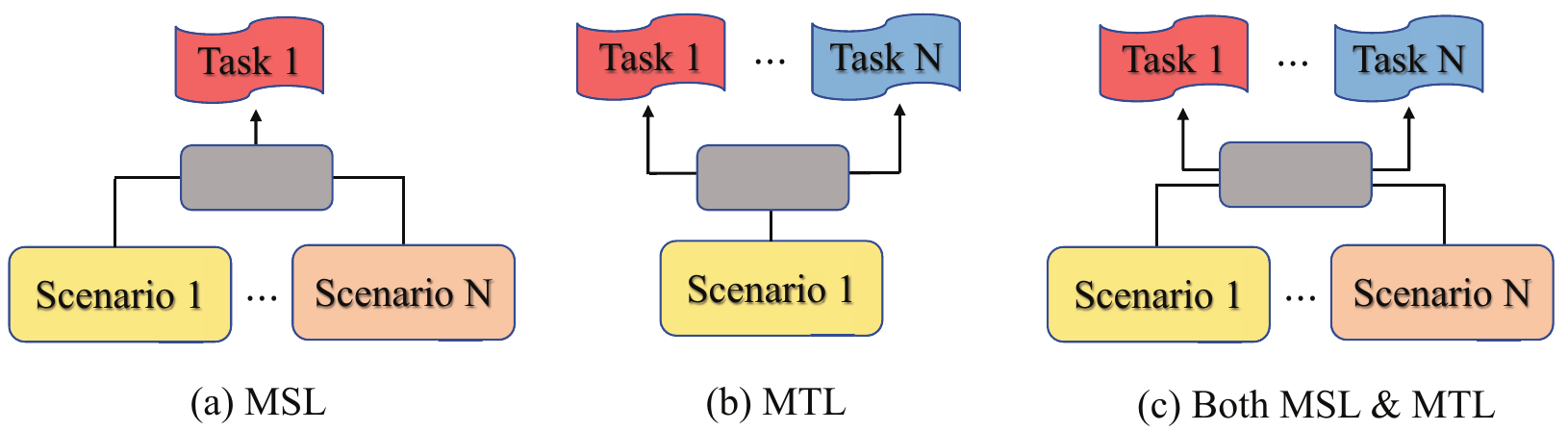}  
\caption{Comparison of MSL, MTL, and MSL\&MTL.}
\label{fig:MSLMTL}
\end{figure}

Naturally, many of these scenarios share some useful information towards user's intent, such as user profile, item features, and even the same queries. Apparently, searching systems can be enriched through information sharing across relevant scenarios with a Multi-scenario learning (MSL) setting, as shown in Figure~\ref{fig:MSLMTL}(a). Under each scenario, user satisfaction and engagement are often quantified by different objectives, \eg Click-Though Rate (CTR) and Conversion Rate (CVR). Therefore, a Multi-task learning (MTL) setting could also be applied to leverage the intrinsic correlation between different tasks, as shown in Figure~\ref{fig:MSLMTL}(b).

Recently, significant efforts have been devoted to devise different MSL and MTL solutions~\cite{caruana1997multitask,MMoE,ESMM,HMoE,Cross-Stitch,PLE,Sluice,STAR,TreeMS}. These solutions mainly exploit the multi-gate mixture-of-experts (MMOE) architecture to distill scenario-/task-specific information as well as the shared information among scenarios and tasks. To support efficient information sharing and alleviate the negative transfer~\cite{torrey2010transfer}, these solutions mainly resort to domain knowledge to design specific expert structure. For example, for MSL, STAR~\cite{STAR} proposes a star topology with a shared center and scenario-specific experts. TreeMS~\cite{TreeMS} chooses to model different E-Commerce scenarios with a tree structure. For MTL, PLE~\cite{PLE} handcrafts  explicit separation between the task-specific and task-shared experts, for video and product recommendations. Note that these strategies require strong domain expertise to optimize the network structure. Nevertheless, the resultant structures are globally static across different scenarios, tasks, and inputs.

It is now becoming a business need to optimize a bunch of tasks across different scenarios, as shown in Figure~\ref{fig:MSLMTL}(c). A straightforward solution is to deploy different MSL models for different tasks. On the one hand, this simple strategy will sacrifice the intrinsic correlations between the tasks that are validated to be beneficial in existing MSL solutions. On the other hand, optimizing the network structure for each task on the basis of MSL requires further expertise, leading to significant increase in maintenance cost. The same story can be told by reversing this order. Here, we are particularly interested in developing a unified framework that could realize both MSL and MTL requirements, and alleviate the above problems.

To this end, in this paper, we propose a novel \textbf{A}utomatic \textbf{E}xpert \textbf{S}election framework for \textbf{M}ulti-scenario and \textbf{M}ulti-task search, named \baby. To the best of our knowledge, this is the first work to tackle both MSL and MTL problems simultaneously within a unified ranking framework. \baby is a flexible hierarchical structure where the multi-task layers are stacked over the multi-scenario layers. This hierarchical design bears two unique merits: 1) we can model the complex relations between different scenarios for more efficient information sharing. 
%As described in Figure xxx, 
In a business setting, many scenarios can be organized as a hierarchical structure in nature. 
The hierarchical modeling of  business logic further enhances efficiency of information sharing. Moreover, this hierarchical structure leads to significant reduction in computational cost (ref. Section~\ref{ssec:ourmodel}); 2) we can make full use of intrinsic correlations among different tasks without deploying the different MSL design under each task, and vice versa. Again, this leads to large saving of maintenance cost.  

Another core novelty in \baby is a general expert selection algorithm for automatically extracting scenario/task-specific and the shared information, in each multi-scenario and multi-task layer. In detail, the selection process leverages scenario/task correlations and is anchored on the sparsification of multi-gate mixture-of-experts structure. 
We emphasize that expert selection is performed for each instance, so as to achieve an adaptive structure learning. We also devise an auxiliary loss function to guide the training of this selection process. 

We perform extensive experiments on two real-world datasets, named AliPay and AliExpress, with different characteristics. Experimental results show that \baby achieves substantial improvements over state-of-the-art models and possible technical alternatives. The online experiments also verify the superiority of our proposed \baby. To summarize, we make the following contributions in this paper:
\begin{itemize}
    \item We introduce a novel automatic expert selection framework for multi-scenario and multi-task search. To the best of our knowledge, \baby is the first work to enable both MSL and MTL simultaneously under a unified ranking framework.
    \item We devise a flexible and efficient hierarchical structure in \baby to accommodate complex relations underlying different scenarios. Moreover, an expert selection algorithm is proposed to support adaptive structure learning, such that task-specific and shared information can be better harnessed in instance level. To guide the selection process, we further introduce an auxiliary loss to enhance model training.
    \item We conduct extensive experiments on two real-world large-scale searching datasets. Both offline and online experiments well demonstrate of the superiority of our proposed \baby. Currently, \baby has been successfully deployed in AliPay to serve major online traffic.
\end{itemize}

%=========================
\section{Related Work}
%=========================

In this section, we briefly review existing studies that are highly related to our work, specifically, multi-scenario learning, multi-task learning, and sparse experts network.

%================================
\subsection{Multi-Scenario Learning}
%================================

Multi-scenario learning aims at addressing the same task with a unified model under different scenarios, where the label spaces are of the same. STAR~\cite{STAR} tries  to capture scenario correlations via star topology which consists of shared center parameters and scenario-specific parameters.  Chen~\etal utilize an auxiliary network to model shared knowledge separately~\cite{SAML}. ZEUS~\cite{ZEUS} focuses on numerous user behaviors across all scenarios, and performs self-supervised learning to improve ranking performance. Since MSL and MTL can be viewed in a unified perspective~\cite{yang2014unified}, other studies~\cite{HMoE, TreeMS} formulate this problem as a multi-task learning problem, and aim to optimize all scenarios with MMoE-like structure in an end-to-end way. Specifically, the core part of a multi-gate mixture-of-experts (MMoE)~\cite{MMoE} is the gating mechanism that fuses the output of different parallel expert networks together. HMoE ~\cite{HMoE} represents each scenario with experts respectively, while TreeMs~\cite{TreeMS} represents scenarios as a semantic tree and utilize LSTM~\cite{LSTM} to model scenario relations. However, all work discussed above are designed to adjust the basic MMoE structure to adapt a certain application.

%==================================
\subsection{Multi-Task Learning}
%==================================

Unlike MSL, multi-task learning focuses on modeling both intrinsic correlations (or information sharing) and specific knowledge across different tasks~\cite{MMoE, PLE, xi2021modeling, zhao2019multiple}, \eg click-though rate prediction \& conversion rate prediction~\cite{ESMM, HMoE} and credit risk forecasting \& credit limits setting~\cite{CRF}. Hard parameter sharing~\cite{caruana1997multitask} is the simplest model to learn intrinsic correlations through a shared bottom layer. To model more complex task correlations, recent work~\cite{Cross-Stitch, Sluice} resort to MMoE to realize information sharing. However, they suffer from ``Seesaw Phenomenon''~\cite{PLE} - the improvement of one task often leads to performance degeneration of some other tasks. Different from static linear combination, Ma~\etal propose a flexible parameter sharing framework in multi-task setting~\cite{SNR}. Recently, PLE~\cite{PLE} adopts a progressive routing mechanism and separates experts into task-sharing and task-specific group in an explicit way, which alleviates the Seesaw Phenomenon in recommender systems.

%==============================
\subsection{Sparse Experts Network}
%==============================

Sparse experts network has been proven to be useful for improving computational efficiency and enhancing model performance~\cite{shazeer2017outrageously, wang2020deep}. Various sparse policies have been studies where sparsity can be achieved through hard setting (\eg setting values to zeros and ones) like TopK~\cite{xiao2021adversarial, yang2021exploring, ruiz2021scaling} or in a more relaxed manner like gumbel-softmax~\cite{nie2021dense, kool2021unbiased}. However, these work only considers sparse policy with regard to a single gate \ie MoE~\cite{MoE}, which is not directly generalizable to multi-task and multi-scenario settings. To the best of our knowledge, our work is the first attempt that adapts sparse policy according to the task/scenario correlations on top of the basic MMoE structure.

%=============================
\section{Method}
\label{sec:method}
%=============================

\begin{figure}[th]
\centering
\includegraphics[height=96mm,width=0.88\linewidth]{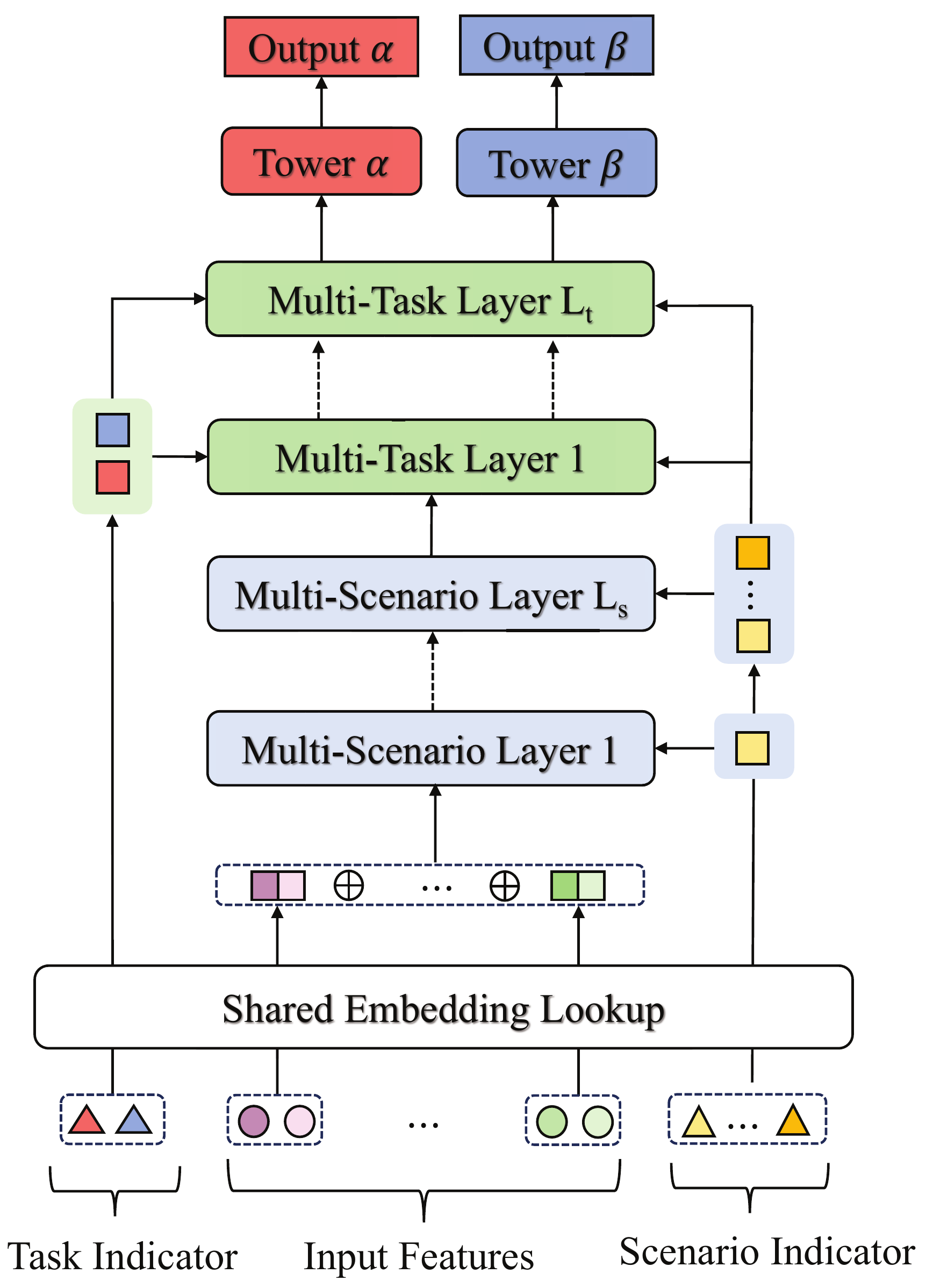}
\caption{The whole architecture of \baby.}\label{fig:overview}
\end{figure}

\begin{figure*}[th]
\centering
\begin{subfigure}[MMoE]{\includegraphics[width=.27\linewidth]{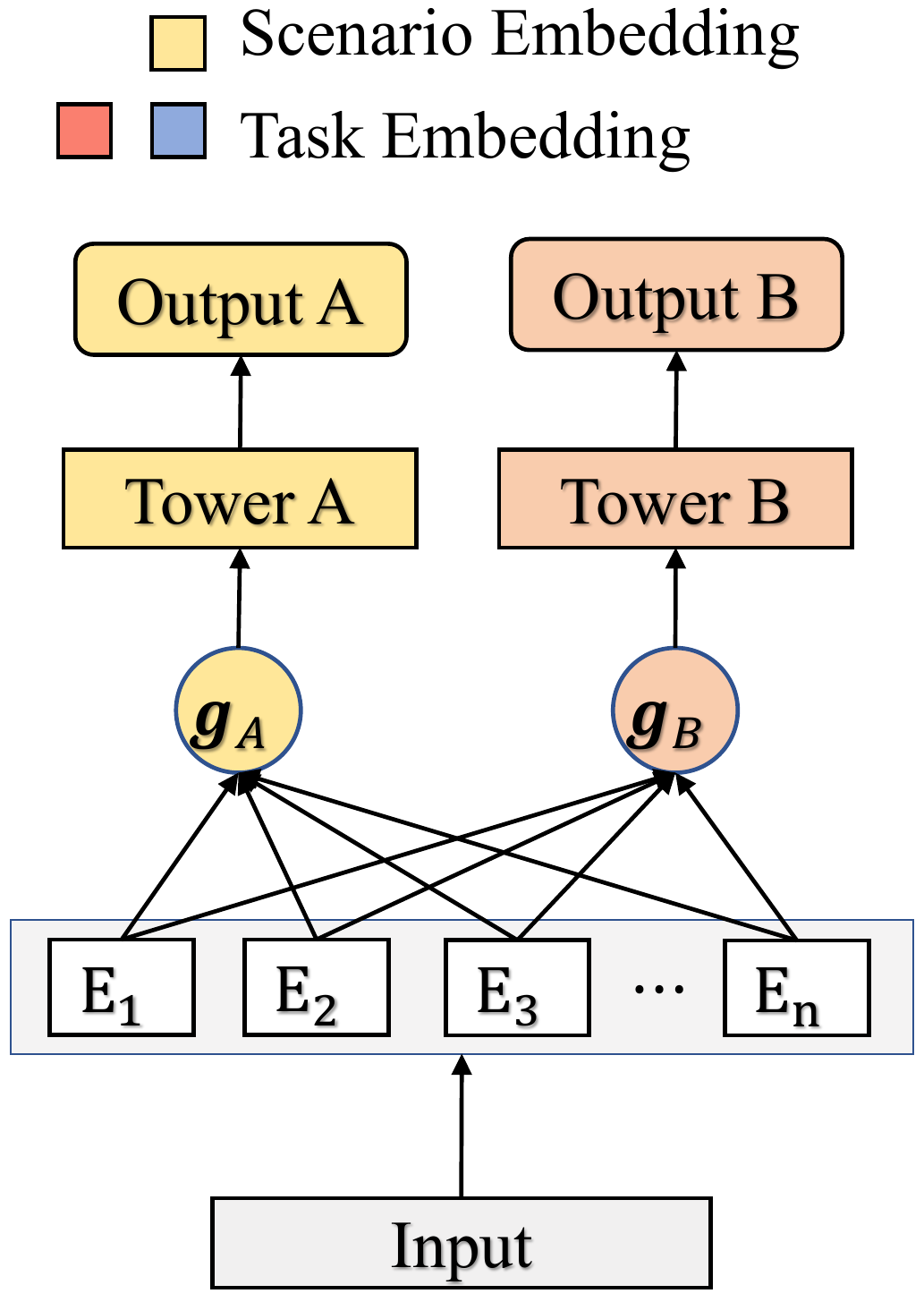}
   \label{fig:MMoE}
 }%
\end{subfigure}\hfill
\begin{subfigure}[Multi-Scenario Layer]{\includegraphics[width=.30\linewidth] {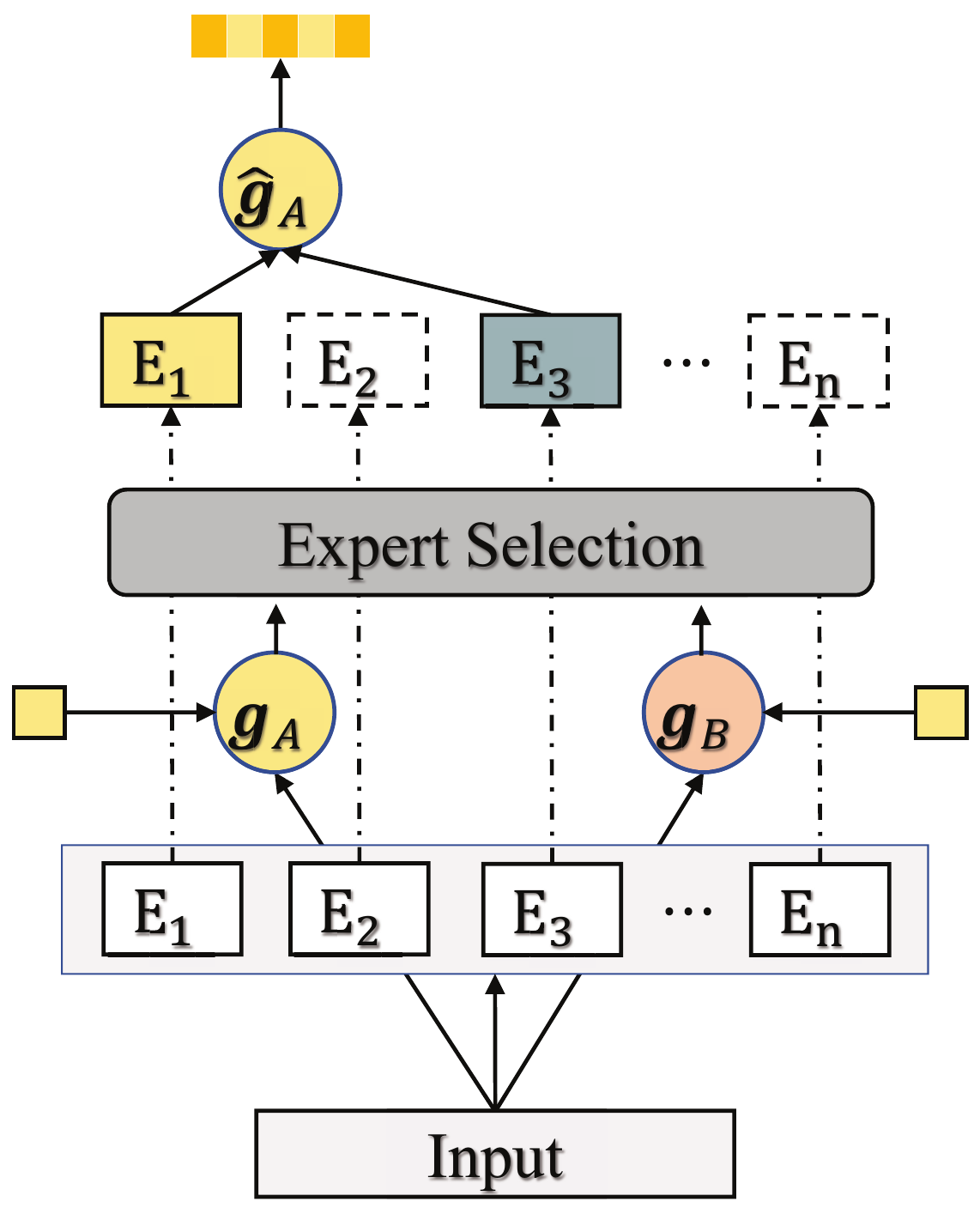}
   \label{fig:MSLayer}
 }%
\end{subfigure}\hfill
\begin{subfigure}[Multi-Task Layer]{\includegraphics[width=.30\linewidth] {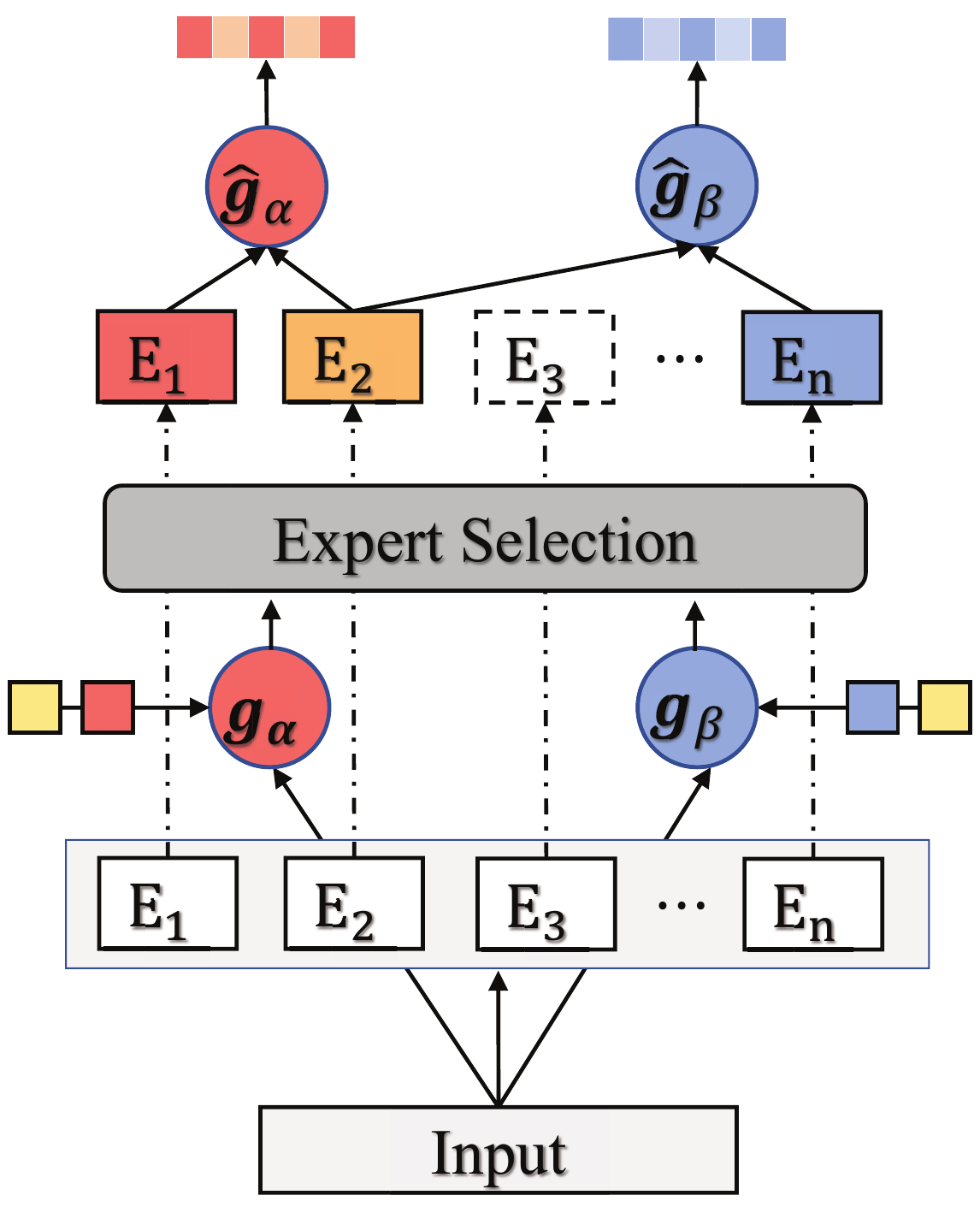}
   \label{fig:MTLayer}
 }%
\end{subfigure}
\caption{(a) Network structure of MMoE. (b) and (c) are illustrations of a single multi-scenario layer and a single multi-task layer, respectively in \baby}
\end{figure*}

In this section, we first formulate our research problem. Then, we introduce the basics of MMoE which serve as preliminary knowledge for our model. After detailing the components in our model, we present model optimization procedures.

%=============================
\subsection{Problem Formulation}
%=============================

In a searching system, users retrieve items through queries. In a multi-scenario \& multi-task searching setting, users may be put under different scenarios for various tasks optimization. Formally, we have $\{\mathcal{S}, \mathcal{T}, \mathcal{U}, \mathcal{I}, \mathcal{Q}\}$. 
Here, $\mathcal{S}$ and $\mathcal{T}$ denote scenario set and task set respectively. User set $\mathcal{U}$ is composed of users from all scenarios: $\mathcal{U} = \mathcal{U}_{1} \cup \mathcal{U}_{2} \cdots \mathcal{U}_{s}$. The item set $\mathcal{I}$ and query set $\mathcal{Q}$ are  defined in the same way. Given extensive features of users, items, and queries, our research problem is to optimize all tasks in $\mathcal{T}$ for every scenario. The goal is to learn a rating function $y_t(u,i,q)$ for each task: $\mathcal{U} \times \mathcal{I} \times \mathcal{Q} \rightarrow \mathbb{R^{\mid \mathcal{S}\mid \times \mid \mathcal{T}\mid }}$, where $y_t(u,i,q)$ is the ranking score of item $i$ for user $u$ who issues query $q$ in the perspective of task $t$.

%=============================
\subsection{Basics of MMoE}
%=============================

The Multi-gate Mixture-of-Experts (MMoE) is designed to tackle multi-task problem, and is widely used in online platforms for ranking~\cite{MMoE, HMoE} or addressing cold-start problem~\cite{TreeMS, zhu2020recommendation}. We  briefly review MMoE's network architecture, illustrated in Figure~\ref{fig:MMoE}. 

Suppose there are $n$ experts and $m$ tasks, and each expert is implemented as independent multi-layer perception, \ie $f_{i}(\cdot)$ and $1\leq i\leq n$. A MMoE combines these experts together through a gating mechanism. Specifically, let $\mathbf{x}$ denote the input vector. The output of this MMoE module for a certain task $t$ is shown as follows:
\begin{align}
o_t(\mathbf{x})&=MMoE(\mathbf{x},\mathbf{g})=\sum_{i=1}^{n} \mathbf{g}_{t}[i]f_i(\mathbf{x}) \label{eqn:MMoE}\\
f_{i}(\mathbf{x})&=\sigma(\mathbf{W}_i\mathbf{x})
\end{align}
where $\sigma(\cdot)$ is the nonlinear activation function, and $\mathbf{g}_{t}\in\mathbb{R}^{n}$ denotes gating scalar vector. Here, $\mathbf{g}_{t}[i]$ indicates the importance score for $i$-th expert, which is calculated as follows:
\begin{align}
\mathbf{g}&=\mathbf{W}_{t} \mathbf{x}\\
\mathbf{g}_{t}&=softmax(\mathbf{g})
\end{align}
where $\mathbf{W}_{t}$ is a learnable parameter of gating mechanism for task $t$.

%=============================
\subsection{The Proposed Model}
\label{ssec:ourmodel}
%=============================

Our model \baby is illustrated in Figure~\ref{fig:overview}. \baby treats multi-scenario and multi-task in a similar way and adopts a hierarchical architecture to couple them into a unified framework. Different from optimizing a static structure, \baby leverages a general and flexible architecture for selecting scenario/task-specific and scenario/task-shared experts automatically. In the following, we first introduce the shared embedding layer in \baby. For simplification, we detail a simple model with only a single multi-scenario layer (Figure~\ref{fig:MSLayer}) and a single multi-task layer (Figure~\ref{fig:MTLayer}). Lastly, we discuss the extension in a multi-layer setting.

\paratitle{Shared Embedding Layer.}
The shared embedding layer is devised to cast the raw categorical and numerical features into continuous embeddings. Supposed there are $v$ groups of features, such as user group (\eg age, gender), item group (\eg price, brand), and query group (\eg query frequency). For a given instance, we first transform numerical features into categorical types, and then encode each categorical feature to one-hot vector. Let user group be the first group, we project these one-hot vectors into corresponding embeddings and form the group embedding $\mathbf{x}_1$ as follows:
\begin{align}
\mathbf{x}_{1}=\left[\mathbf{W}_{1}^{u} \mathbf{x}_{1}^{u},\mathbf{W}_{2}^{u} \mathbf{x}_{2}^{u}, \ldots, \mathbf{W}_{n_{u}}^{u} \mathbf{x}_{n_{u}}^{u} \right]
\label{eqn:xu}
\end{align}
where $n_{u}$ is the number of features in user group, $[\cdot,\cdot]$ refers to vector concatenation. The same applies to other groups. Then we concatenate embeddings of all groups together to form the feature vector $\mathbf{x}$ as the input to the subsequent components in \baby:
\begin{equation}
\mathbf{x}=\left[\mathbf{x}_{1},\mathbf{x}_{2},\cdots,\mathbf{x}_{v}\right]
\end{equation}
where $\mathbf{x}\in \mathbb{R}^{d}$ and $d$ is dimension size.

Meanwhile, scenario embedding $\mathbf{s}$ that indicates the scenario of a given data record, and task embeddings $\mathbf{t}_{1}$, $\mathbf{t}_{2}$,  $\cdots$ , $\mathbf{t}_{\mid\mathcal{T}\mid}$ including all tasks, are generated from scenario or task ids respectively. $\mid\mathcal{T}\mid$ is the size of task set.

\paratitle{Multi-Scenario Layer.} After the shared embedding layer, we feed $\mathbf{x}$ into the multi-scenario layer for expert selection, to get the corresponding output in terms of the selected experts. For simplicity, we only show the selecting process of one layer in Figure~\ref{fig:MSLayer}, multiple layers can be stacked together if needed.

Suppose there are $n$ experts and $m$ scenarios (\ie $m$ gates) in a multi-scenario layer. We first utilize a linear transformation to calculate the gating scalar vector for each scenario as follows:
\begin{align}
\mathbf{G}&=[\mathbf{g}_1,\cdots,\mathbf{g}_m]\\
\mathbf{g}_j&=\mathbf{S}_{j}[\mathbf{x},\mathbf{s}] + \bm{\eta}_{j}
\label{eqn:gs}
\end{align}
Here $\mathbf{g}_{j}[k]$ is the relevance score between the $j$-th scenario and the $k$-th expert with respect to the input instance ($\mathbf{g}_j\in\mathbb{R}^n$). $\mathbf{S}_j$ is a learnable parameter;  $\mathbf{s}$ is the scenario embedding of the given data record; $\bm{\eta}_{j}$ is the Gaussian noise whose scale is positively proportional to $n$ (\ie $ \|\bm{\eta}_{j}\| \propto n$); and $\mathbf{G}\in\mathbb{R}^{n\times m}$. The introduced gaussian noise will guide our model to explore search space for better expert selection.

Expert row vector $\mathbf{G}[k,:] = \big[\mathbf{G}[k,1],\cdots,\mathbf{G}[k,m]\big]$ is the key for expert selection. We assume that if $\mathbf{G}[k,j]$ gets relatively higher score than the other elements in $\mathbf{G}[k,:]$, the $k$-th expert is more likely classified into scenario-specific group of $j$-th scenario.  If values in $\mathbf{G}[k,:]$ all get closer to each other relatively, then the $k$-th expert may contain shared information among all scenarios. Hence, we firstly perform a row-wise softmax operation over matrix $\mathbf{G}$ as follows:
\begin{align}
\mathbf{\tilde{G}}=softmax(\mathbf{G})\label{eqn:tildeG}
\end{align}
Then, for the $j$-th scenario, we form an one-hot scenario indicator vector $\mathbf{p}_j\in\mathbb{R}^m$ where only the $j$-th element is $1$ (\eg $[1,\cdots,0]$ for the first scenario). As aforementioned, the likelihood of the $k$-th expert being a scenario-specific group for $j$-th scenario can be measured by the similarity between $\mathbf{p}_j$ and $\mathbf{\tilde{G}}[k,:]$. 

Similarly, we also introduce an uniform vector $\mathbf{q}_j\in\mathbb{R}^m$ for $j$-th scenario: $\mathbf{q}_j=[1/m,\cdots,1/m]$. The likelihood of $k$-th expert being a scenario-shared group for $j$-th scenario can be measured by the similarity between $\mathbf{q}_j$ and $\mathbf{\tilde{G}}[k,:]$. Formally, the selection process for generating scenario-specific and scenario-shared expert indices is described as follows:
\begin{align}
\mathcal{E}_{sp} &= TopK(h^p_1,\cdots,h^p_n)\\
h_k^p&=- KL(\mathbf{p}_{j}, \mathbf{\tilde{G}}[k,:])\\
\mathcal{E}_{sh} &= TopK(h^q_1,\cdots,h^q_n)\\
h_k^q&=- KL(\mathbf{q}_{j}, \mathbf{\tilde{G}}[k,:])
\end{align}
where $\mathcal{E}_{sp}$ and $\mathcal{E}_{sh}$ respectively refer to the scenario-specific and scenario-shared expert index set for $j$-th scenario. Function $TopK(\cdot)$ is the argument operation that returns the indices of top-$K$ values, and $KL(\cdot)$ calculates the Kullback-Leibler divergence where a lower divergence score means a higher similarity of the two distributions.

With $\mathcal{E}_{sp}$ and $\mathcal{E}_{sh}$, we update the gating scalar vector $\mathbf{g}_{j}$ to accomplish sparse experts network:
\begin{align}
\label{eqn:hatg}
   \mathbf{\hat{g}}_j[k]=
   \begin{cases}
        \quad\mathbf{g}_{j}[k], & \textit{if} \quad k \in \mathcal{E}_{sh}\cup\mathcal{E}_{sp} \\
        \quad-\infty, & \textit{else} \\
\end{cases}
\end{align}
Following Equation~\ref{eqn:MMoE}, the output of this scenario layer can be derived with the updated gating scalar matrix $\mathbf{\hat{g}}_{j}$ and the outputs of $n$ experts via a standard MMoE module:
\begin{align}
\mathbf{z}_{j} = ScenarioLayer(\mathbf{x},\mathbf{s}_j) = MMoE(\mathbf{\mathbf{x},\hat{g}}_{j})
\end{align}
where $\mathbf{z}_j$ is the output for $j$-th scenario.

\paratitle{Multi-Task Layer.} Given the output $\mathbf{z}_j$ from the above multi-scenario layer, we also utilize the same selection process to achieve multi-task learning. Assuming there are $n$ experts and $t$ tasks in the multi-task layer, we calculate the gating scalar vector $\mathbf{g}_k$ for $k$-th task as follows:
\begin{align}
\mathbf{g}_k&=\mathbf{T}_{k}[\mathbf{x},\mathbf{s},\mathbf{t}_k] + \bm{\eta}_{k}
\label{eqn:gj}
\end{align}
where $\mathbf{T}_k$ is a learnable parameter, $\mathbf{t}_k$ denotes the embedding of $k$-th task, $\bm{\eta}_t$ is the Gaussian noise similar to Equation~\ref{eqn:gs}. By forming the one-hot task indicator vector and uniform vector for each task, we select the task-specific experts for each task and task-shared experts respectively, following Equations~\ref{eqn:tildeG}-\ref{eqn:hatg}. The resultant gating scalar vector $\mathbf{\hat{g}}_{k}$ is used to derive the output of each task via a MMoE module:
\begin{align}
\mathbf{z}_k=TaskLayer(\mathbf{z}_j,\mathbf{t}_k)=MMoE(\mathbf{\mathbf{z}_j,\hat{g}}_{k})
\label{eqn:tasklayer}
\end{align}
Then, we feed the output $\mathbf{z}_k$ into a tower of MLP to generate the prediction for $k$-th task:
\begin{align}
\hat{y}_k=\sigma(MLP(\mathbf{z}_k))
\label{eqn:yj}
\end{align}
where $\hat{y}_k$ is the prediction for $k$-th task.

\begin{figure}[!]
\centering  
\includegraphics[width=0.42\textwidth]{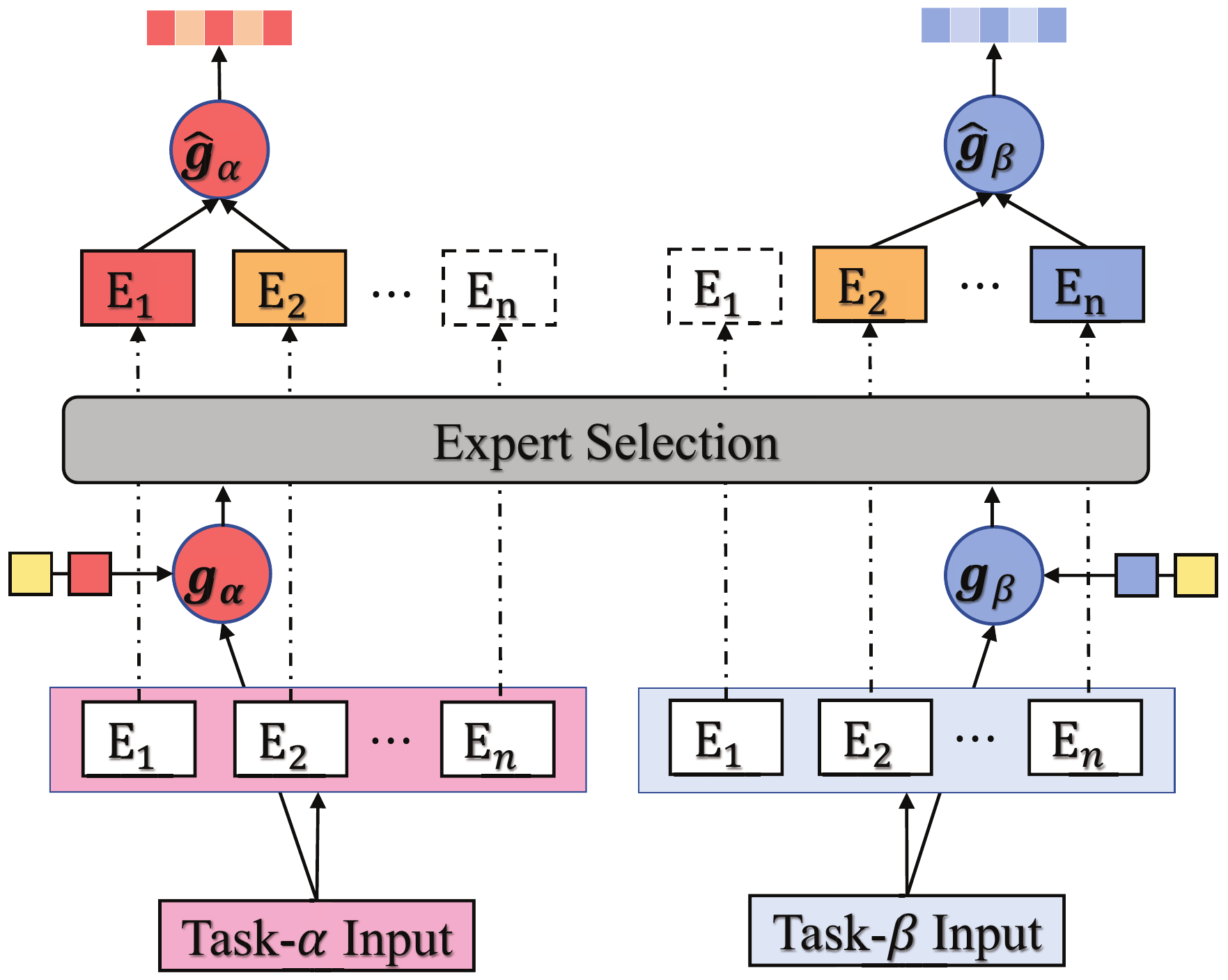}  
\caption{The network structure of multi-task layers with different task inputs.}
\label{eqn:highMTLayer}  
\end{figure}

\paratitle{Multi-Layer Extension.}
In real-world applications, a scenario could be complex and exhibits a hierarchical structure in nature~\cite{TreeMS, xiao2021adversarial}. 
%As shown in Figure~xxx, 
For example, two search services $\alpha_1$ and $\alpha_2$ for two different domains (\eg search by brand, and search by feature) can be accessed through two different channels $\beta_1$ and $\beta_2$ (\eg home page and vertical page shown in Figure~\ref{fig:intro}). These two channels may hold some unique characteristics against each other, and could be considered as two parent scenarios. Hence, we can form four fine-grained scenarios: $\beta_1\to\alpha_1$, $\beta_1\to\alpha_2$, $\beta_2\to\alpha_1$ and $\beta_2\to\alpha_2$. 

Our proposed \baby can be easily extended to accommodate this complex situation by stacking multiple multi-scenario layers together (each with a different parameter set). As shown in Figure~\ref{fig:overview}, we can stack multiple multi-scenario layers to reach more complex multi-scenario modeling. Given there are $L_s$ multi-scenario layers, the only modification in $l$-th layer is to update the scenario embedding $\mathbf{s}$ in Equation~\ref{eqn:gs} as $[\mathbf{s}_1,\cdots,\mathbf{s}_l]$, where $s_i$ is the embedding of the specified scenario in $i$-th layer, $1<l\le L_s$. In fact, we can model $n_1\times n_2\cdots n_{L_s}$ fine-grained scenarios. Note that when if a standard MMoE architecture is used, we need the same number of experts. In this sense, we can significantly reduce computation cost because $\sum_{i=1}^{L_s}n_i\ll n_1\times n_2\cdots n_{L_s}$.

\begin{table*}[!]
  \caption{Statistics of AliPay dataset. $\#impress$, $\#clk$, and $\#cv$ are the numbers of impressions, clicks, conversions respectively. $all$ is the merged data of all scenarios. $\#impress\%$ represents the percentage of $\#impress$ in one scenario among all scenarios, similarly  $\#clk\%$, and $\#cv\%$.}
  \label{tab:alipay_stat}
  \begin{tabular}{c|r|rrrr}
    \toprule
    \ &ALL& HP\&RI &HP\&BS  & VP\&RI & VP\&BS\\
    \midrule
    \#impress & 96,612,494 & 12,543,990  & 27,930,226 &49,919,544 & 6,218,734\\
    \#clk  & 26,519,344  & 3,554,662  & 650,338 & 21,833,316 & 481,028 \\
    \#cv   & 1,664,640  &310,917  &25,495  &1,311,422  &16,806 \\
    \midrule
    \#clk/\#impress  &27.45\%  &28.34\%  &2.33\%  &43.74\%  &7.74\% \\
    \#cv/\#impress  &1.72\%  &2.48\%  &0.09\%  &2.63\%  &0.27\% \\
    \#cv/\#clk  &6.28\%  &8.75\%  &3.92\%  &6.00\%  &3.49\% \\
    \midrule
    \#impress\% &100\%  &12.98\%  &28.91\%  &51.67\%   &6.44\% \\
    \#clk\% &100\%  &13.40\%  &2.45\%  &82.33\%   &1.82\% \\
    \#cv\% &100\%  &18.68\%  &1.53\%  &78.78\%   &1.00\% \\
  \bottomrule
\end{tabular}
\end{table*}

\begin{table*}[!]
  \caption{Statistics of AliExpress dataset.}
  \label{tab:ae_stat}
  \begin{tabular}{l|r|rrrr}
    \toprule
    \ &ALL & NL & FR & ES & US\\
    \midrule
    \#impress & 103,814,836  & 17,717,195 &27,035,601  &31,669,427  &27,392,613\\
    \#clk &2,215,494  &381,078  &542,753  &842,055  &449,608 \\
    \#cv & 58,171  &13,815  &14,430  &19,096  &10,830 \\
    \midrule
    \#clk/\#impress  &2.13\%  &2.15\%  &2.00\%  &2.66\%  &1.64\% \\
    \#cv/\#impress  &0.06\%  &0.08\%  &0.05\%  &0.06\%  &0.04\% \\
    \#cv/\#clk  &2.63\%  &3.63\%  &2.66\%  &2.27\%  &2.41\% \\
    \midrule
    \#impress\% &100\%  &17.07\%  &26.04\%  &30.51\%   &26.39\% \\
    \#clk\% &100\%  &17.20\%  &24.50\%  &38.01\%   &20.29\% \\
    \#cv\% &100\%  &23.75\%  &24.81\%  &32.83\%   &18.62\% \\
  \bottomrule
\end{tabular}
\end{table*}

As the design of \baby is modular and flexibly, we can stack $L_t$ multi-task layers together for higher-level feature extraction for each task (ref. Figure~\ref{eqn:highMTLayer}). However, things become slightly  different because we have different input for each task after the first multi-task layer. To address this, we decide to share the same set of experts in each of the later multi-task layers, and perform expert selection with different inputs. Following Equations~\ref{eqn:gj} and~\ref{eqn:tasklayer}, we derive the output for $l$-th multi-task layer by taking the output from the preceding layer $\mathbf{z}_k^{l-1}$ as input, where $l>1$. Also, the scenario embedding $\mathbf{s}_j$ used in Equation~\ref{eqn:gj} keeps the same as $[\mathbf{s}_1,\cdots,\mathbf{s}_{L_s}]$ for all multi-task layers. Figure~\ref{eqn:highMTLayer} illustrates the architecture of such a multi-task layer. The prediction is made in terms of the output of the last multi-task layer by following Equation~\ref{eqn:yj}.

\begin{align}
\mathcal{L}_{k}=-y_{j}^{k} \log \left(\hat{y}_{j}^{k}\right)-\left(1-y_{j}^{k}\right) \log \left(1-\hat{y}_{j}^{k}\right)
\end{align}
where ${y}_{j}^{k}$ is the ground-truth label of $j$-th instance on $k$-th task, \eg a click (\ie $CTR$) or a conversion (\ie $CTCVR$), and $\mathcal{L}_k$ is the prediction loss for $k$-th task.

%=============================
\subsection{Model Optimization}
%=============================

Since the prediction for search task can be formulated as a probability prediction problem, we choose sigmoid function as the activation in Equation~\ref{eqn:yj}. Accordingly, we adopt the widely used Cross-Entropy loss for $j$-th instance as follows:

\begin{table*}[!]
  \caption{Performance comparison in AliPay dataset. The best results are in boldface and second best underlined. Symbol * indicates that difference to the best baseline is statistically significant at 0.05 level.}
  \label{tab:res_alipay}
\begin{tabular}{c||cc|cc|cc|cc|cc}
\toprule
\multirow{2}{*}{Model} & \multicolumn{2}{c|}{HP\&RI} & \multicolumn{2}{c|}{HP\&BS} & \multicolumn{2}{c|}{VP\&RI} & \multicolumn{2}{c|}{VP\&BS} & \multicolumn{2}{c}{ALL} \\ \cmidrule{2-11} 
                       & auc\_ctr  & auc\_ctcvr & auc\_ctr  & auc\_ctcvr & auc\_ctr  & auc\_ctcvr & auc\_ctr  & auc\_ctcvr & auc\_ctr  & auc\_ctcvr  \\
\midrule
Hard Sharing & 90.89 & 90.95 & 77.14 & 81.60 & 86.27 & 85.56 & 69.51  & 75.64 & 90.34 & 89.15     \\
Parallel DNN & \underline{91.22} & 91.55 & 77.03 & 81.37 & 86.34 & 85.79 & 70.57  & 76.46 & 90.62 & 89.60     \\
Cross-Stitch & 91.12 & 91.59 & 77.06 & 81.82 & \underline{86.39} & 86.08 & 70.86  & 76.79 & \underline{90.63} & 89.82     \\
STAR         & 90.62 & 91.66 & 74.04 & 81.86 & 85.66 & 86.34 & 69.29  & 77.13 & 89.98 & 90.00    \\
MMoE         & 91.06 & 91.71 & \underline{77.54} & \underline{82.31} & 86.36 & 86.36 & 69.94  & \underline{77.18} & 90.62 & 90.03     \\
HMoE         & 89.05 & 89.54 & 75.61 & 79.68 & 85.32 & 85.24 & 65.52  & 70.56 & 89.58 & 88.72     \\
PLE          & 91.11 & \underline{91.76} & 77.05 & 81.93 & 86.35 & \underline{86.47} & \underline{70.90}  & 77.03 & \underline{90.63} & \underline{90.10}     \\
\baby          & \textbf{91.49}$^{*}$ & \textbf{ 92.21}$^{*}$ & \textbf{ 78.22}$^{*}$ & \textbf{ 82.98}$^{*}$ & \textbf{ 86.73}$^{*}$ & \textbf{ 86.91}$^{*}$ & \textbf{ 72.13}$^{*}$  & \textbf{ 78.33}$^{*}$ & \textbf{ 90.86}$^{*}$ & \textbf{ 90.42}$^{*}$ \\
\bottomrule
\end{tabular}
\end{table*}

We also propose an auxiliary loss $\mathcal{L}_{aux}$ to guide the scenario/task expert selection. Specifically, for $L_s$ multi-scenario layers, we aim to enhance the expert selection decision during the training process as follows:
\begin{align}
\mathcal{L}_{sp}^{s} =\sum_{l=1}^{L_s}\sum_{k\in\mathcal{E}_{sp}}KL(\mathbf{p}_{j},\mathbf{\tilde{G}}^l[k,:])
\label{eqn:loss_sp}
\end{align}
where $\mathbf{\tilde{G}}^l[k,:]$ is the output of Equation~\ref{eqn:tildeG} in $l$-th multi-scenario layer. Similarly, we can enhance the expert selection decision for $L_t$ multi-task layers with loss $\mathcal{L}_{sp}^t$. The two auxiliary losses are summed as $\mathcal{L}_{sp}=\mathcal{L}_{sp}^{s} + \mathcal{L}_{sp}^{t}$. The same goes for the expert selection in the scenario/task-shared side, denoted by $\mathcal{L}_{sh}$. The final auxiliary loss $\mathcal{L}_{aux}$ is then defined as follows:
\begin{align}
\mathcal{L}_{aux} = \lambda_{sp}\mathcal{L}_{sp} + \lambda_{sh}\mathcal{L}_{sh}
\label{eqn:loss_Laux}
\end{align}
where $\lambda_{sp}$ and $\lambda_{sh}$ are hyper-parameters
to tune. That is, $\mathcal{L}_{sp}$ enhances the selecting process by guiding the gating scalar distribution of a specific experts network get close to the assumed specific distribution $\mathbf{p}_j$. The same goes for $\mathcal{L}_{sh}$. Finally, we get the overall loss function for our model:
\begin{equation}
\mathcal{L}=\sum_{k=1}^t\lambda_{k}\mathcal{L}_k+ \mathcal{L}_{aux}+ \gamma\|\Theta\|_{2}^{2}
\label{eqn:loss_L}
\end{equation}
where $\lambda_{k}$ and $\gamma$ are hyper-parameters, and $\|\Theta\|_{2}$ denotes $L2$ regularization over model parameters. Adam~\cite{Adam} is chosen as the optimizer to perform gradient backpropagation for model optimization.

%=============================
\section{EXPERIMENTS}
\label{sec:exp}
%=============================
In this section, we evaluate our proposed \baby against a series of state-of-the-art baselines. Extensive experiments on two real-world large-scale searching datasets with both multi-scenario and multi-task settings demonstrate the effectiveness of our model, which is further confirmed by the online A/B test across multiple business metrics. At last, we provide in-depth analysis about model components.

\begin{table*}[!]
  \caption{Performance comparison in AliExpress dataset. The best results are in boldface and second best underlined. Symbol * indicates that difference to the best baseline is statistically significant at 0.05 level.}
  \label{tab:res_ae}
\begin{tabular}{c||cc|cc|cc|cc|cc}
\toprule
\multirow{2}{*}{Model} & \multicolumn{2}{c|}{NL} & \multicolumn{2}{c|}{FR} & \multicolumn{2}{c|}{ES} & \multicolumn{2}{c|}{US} & \multicolumn{2}{c}{ALL} \\ \cmidrule{2-11} 
                       & auc\_ctr  & auc\_ctcvr & auc\_ctr  & auc\_ctcvr & auc\_ctr  & auc\_ctcvr & auc\_ctr  & auc\_ctcvr & auc\_ctr  & auc\_ctcvr  \\
\midrule
Hard Sharing & 70.70 & 81.49 & 70.89 & 81.41 & 70.57 & 83.89 & 69.71  & 79.36 & 70.59 & 81.56     \\
Parallel DNN & 71.11 & 84.17 & 70.96 & 84.91 & 70.60 & 85.91 & 69.17  & 79.38 & 70.38 & 83.98     \\
Cross-Stitch & 72.04 & 84.92 & 71.59 & 86.50 & 71.77 & 87.56 & 69.66  & 83.02 & 71.28 & 85.91     \\
STAR         & 71.45 & 84.76 & 71.29 & 86.36 & 71.94 & 88.21 & \underline{70.03}  & 85.63 & 71.32 & 86.49     \\
MMoE         & 71.77 & 84.87 & 71.45 & 86.18 & 72.16 & 87.75 & 69.83  & 84.87 & 71.42 & 86.15     \\
HMoE         & 71.87 & 85.47 & 71.56 & 86.76 & 72.06 & 88.01 & 69.81  & 84.71 & 71.41 & 86.50     \\
PLE          & \underline{71.91} & \underline{85.60} & \underline{71.60} & \underline{86.95} & \underline{72.31} & \underline{88.53} & 69.90  & \underline{86.18} & \underline{71.56} & \underline{87.11}     \\
\baby          & \textbf{ 72.60}$^{*}$ & \textbf{ 86.38}$^{*}$ & \textbf{ 72.41}$^{*}$ & \textbf{ 88.08}$^{*}$ & \textbf{ 72.95}$^{*}$ & \textbf{ 89.49}$^{*}$ & \textbf{ 70.88}$^{*}$  & \textbf{ 87.74}$^{*}$ & \textbf{ 72.30}$^{*}$ & \textbf{ 88.12}$^{*}$ \\
\bottomrule
\end{tabular}
\end{table*}

%=============================
\subsection{Experimental Setup}
\label{ssec:expSetup}
%=============================
\paratitle{Datasets.} We collect \textbf{AliPay} dataset sampled from anonymous user daily search logs on AliPay mobile application\footnote{\url{https://global.alipay.com/platform/site/ihome}} from Nov. 15 to Nov. 30 2021. In AliPay dataset, there exists a hierarchical structure among scenarios: $channel \to domain$. There are two channels: home page (HP) and vertical page (VP), and two domains: regular items (RI) and best sellers (BS). As the result, we have four scenarios listed in Table~\ref{tab:alipay_stat},  \eg HP$\to$ RI, VP$\to$RI. We split the data by date for model training and evaluation:  data in Nov. 15-Nov. 28 for training, Nov. 29, 2021 for validation, and Nov. 30, 2021 for testing. Also, there are two tasks for this dataset: click ratio (CTR) and conversion ratio (CTCVR).

Moreover, we also conduct experiments on a public E-Commerce dataset \textbf{AliExpress}\footnote{\url{https://tianchi.aliyun.com/dataset/dataDetail?dataId=74690}}. This dataset also includes multi-scenarios (split by user nationality) and multi-task (CTR and CTCVR) settings. Here, we choose four scenarios: $NL$, $FR$, $ES$, and $US$\footnote{The four scenarios indicate the users from Netherlands (NL), France (FR), Spain (ES) and United States (US).} on AliExpress dataset. Because the original dataset only contains training set and testing set, we randomly sample 50\% of the original testing data as validation set. 

The detailed statistics of the two datasets are reported in Table~\ref{tab:alipay_stat} and~\ref{tab:ae_stat}. Observe that data distribution in both datasets is imbalanced. For example, the number of impressions in VP$\to$BS scenario only accounts for 6.44\% on all scenarios, and majority clicks happen in scenario VP$\to$RI, which accounts for 82.33\% instead.

\paratitle{Baselines.} We compare \baby with two categories of baselines: (1) \textit{gate-aware models} which use gating mechanism to tackle multi scenario or multi-task problems, including MMoE~\cite{MMoE}, HMoE~\cite{HMoE}, and PLE~\cite{PLE}. (2) \textit{gate-free models} including Hard Sharing~\cite{caruana1997multitask}, Parallel DNN, Cross-Stitch~\cite{Cross-Stitch} and STAR~\cite{STAR}. Baselines in the latter category leverage other solutions to model correlations across scenarios or tasks.

\begin{itemize}[leftmargin=1.2em]
    \item \textbf{Hard Sharing}~\cite{caruana1997multitask} is a simple but widely-used model, to encode common knowledge via a shared bottom layer.
    \item \textbf{Parallel DNN} is adapted from basic DNN to fit multiple scenarios or tasks, where  parallel transformation is added for each task/scenario.
    \item \textbf{Cross-Stitch}~\cite{Cross-Stitch} combines multiple networks together via linear cross-stitch units to learn a task-specific representation.
    \item \textbf{MMoE}~\cite{MMoE} exploits multi-gate mixture experts to model relationships among experts implicitly. The merged representations from multiple gates can be transformed to multiple scenario/task prediction layers respectively.
    \item \textbf{HMoE}~\cite{HMoE} extends MMoE to scenario-aware experts with gradient-cutting trick for encoding scenario correlation explicitly. HMoE leverages two models with separate parameters for optimizing CTR and CVR respectively.
    \item \textbf{STAR}~\cite{STAR} adopts star topology with a shared center network and scenario-specific network for CTR prediction task.
    \item \textbf{PLE}~\cite{PLE} is another state-of-art MMoE variant that separates experts into task-specific groups and task-shared groups for the purpose of avoiding negative transfer and seesaw phenomenon.
\end{itemize} 

\paratitle{Metrics and Implementation\footnote{The code implementation is available at \url{https://github.com/alipay/aes_msl_mtl}.}.} Following ESMM~\cite{ESMM}, we optimize CTR and CVR in an entire space, \ie $CTCVR = CTR \times CVR$, where $CTR$ and $CVR$ are the output of CTR tower and CVR tower respectively. The well-known area under ROC curve (AUC) is adopted to evaluate both CTR and CTCVR performances. Data of all scenarios are merged together at input level, and performance under every single scenario as well as the merged all are compared.

Note that, none of baselines consider optimizing MSL\&MTL problem simultaneously within one model. Hence, we extend their structure to adapt to our experiment setting. For example, for multi-scenario model STAR, we implement another star topology in task-level structure. For multi-task model PLE, we extend it with additional progressive layered extraction to tackle multi-scenario problem. In addition, all baselines are trained with merged scenarios data, and are then used to predict both CTR and CTCVR, the same as \baby. Experts in every MMoE-like models are implemented as a single-layer network. For fair comparison, we implement all models with same depth level, all MMoE variant models share the same number of activated experts during training.
%, \ie participate in experts combination via gate mechanism)
PLE and \baby share the same size of specific/shared experts set (\ie $3$ in AliPay and $10$/$3$ for domain/task layer in AliExpress). The $K$ value is set to $1$ and $2/1$ for them respectively.

Recall that, there are two channels and two domains in AliPay dataset before the specific tasks. we implement all models in a stacked way corresponding to the scenario hierarchy of the dataset.

\begin{table}[!]
  \caption{AUC results (CTR) of transfering to different scenarios. The value of row $i$ and column $j$ means trained on $scenario \ i$ and tested on $scenario \ j$}
  \label{tab:SSL_ctr}
  \begin{tabular}{c|cccc}
    \toprule
    \diagbox{Train}{Test} &HP\&RI & HP\&BS & VP\&RI & VP\&BS \\
    \midrule
    \ HP\&RI & 90.61  & 58.68 &76.34  &58.46  \\
    \ HP\&BS & 54.25  & 77.92  &53.28  &63.52  \\
    \ VP\&RI & 86.87& 52.78  &85.85  &55.96 \\
    \ VP\&BS &59.61  &64.91  &60.52  &70.43 \\
  \bottomrule
\end{tabular}
\end{table}

%=============================
\subsection{Performance Comparison}
\label{ssec:expPerformance}
%=============================
Table~\ref{tab:res_alipay} and~\ref{tab:res_ae} report performance of all models on AliPay and AliExpress datasets, respectively.

Results in both tables show that our proposed model \baby  consistently outperforms all baselines  over all scenarios on all tasks. This result  indicates that our model performs well in Multi-scenario\&Multi-task settings and is able to adapt to different configurations of scenarios and tasks. Observe that our model gains significant improvement especially on hard scenarios, \eg VP\&BS scenario of AliPay that suffers from the the highest data sparsity with only 6.44\% impressions.

The results show a broader "SeeSaw Phenomenon" in Multi-scenario\&Multi-task settings among baseline models. That is, multiple tasks could not be improved simultaneously compared to other models in one certain scenario. In other words, performance gain of one metric cannot be achieved consistently across other scenarios. For instance, when compared by CTR in AliPay, PLE performs better than MMoE in VP\&BS scenario but worse than MMoE in HP\&BS scenario. When compared in the scenario of ES in AliExpress dataset, STAR outperforms HMoE on CTR but is poorer than HMoE on CTCVR. This observation is consistent with what has been previously reported in~\cite{STAR,PLE}. However, \baby does not demonstrate SeeSaw phenomenon, compared with other baselines.  

Finally, HMoE performs the poorest, even worse than Hard Sharing on AliPay dataset across all scenarios and tasks, but better performance is observed on AliExpress. One possible reason is that the task layer is attached with deeper levels for AliPay ($channel \to domain \to task$) than AliExpress ($domain \to task$). The capability of modeling scenario correlations for HMoE could be easily corrupted because of the gradient-cutting trick.

\begin{table*}[!]
\small
  \caption{Performance comparison for \baby and its two variants on AliPay dataset. noise: Gaussian noise $\bm{\eta}$ (ref. Equation \ref{eqn:gj} ); auxloss: auxiliary loss $\mathcal{L}_{aux}$ (ref. Equation \ref{eqn:loss_Laux})}
  \label{tab:aba}
\begin{tabular}{c||cc|cc|cc|cc|cc}
\toprule
\multirow{2}{*}{Model} & \multicolumn{2}{c|}{HP\&RI} & \multicolumn{2}{c|}{HP\&BS} & \multicolumn{2}{c|}{VP\&RI} & \multicolumn{2}{c|}{VP\&BS} & \multicolumn{2}{c}{ALL} \\ \cline{2-11} 
                       & auc\_ctr  & auc\_ctcvr & auc\_ctr  & auc\_ctcvr & auc\_ctr  & auc\_ctcvr & auc\_ctr  & auc\_ctcvr & auc\_ctr  & auc\_ctcvr  \\
\hline
\baby w/o (noise\&auxloss)  & 91.30 & 91.91 & 77.92 & 82.63 & 86.48 & 86.44 & 71.72  & 77.41 & 90.71 & 90.02 \\
\baby w/o auxloss           & 91.32 & 91.95 & 78.14 & 82.69 & 86.59 & 86.61 & 71.86  & 77.68 & 90.80 & 90.28 \\
\baby                       & 91.49 & 92.21 & 78.22 & 82.98 & 86.73 & 86.91 & 72.18  & 78.33 & 90.86 & 90.42 \\

\bottomrule
\end{tabular}
\end{table*}

%=============================
\subsection{Analysis of Scenario Relationships}
\label{ssec:expAnalySenario}
%=============================
\begin{figure}[!]  
\centering  
\includegraphics[width=0.48\textwidth]{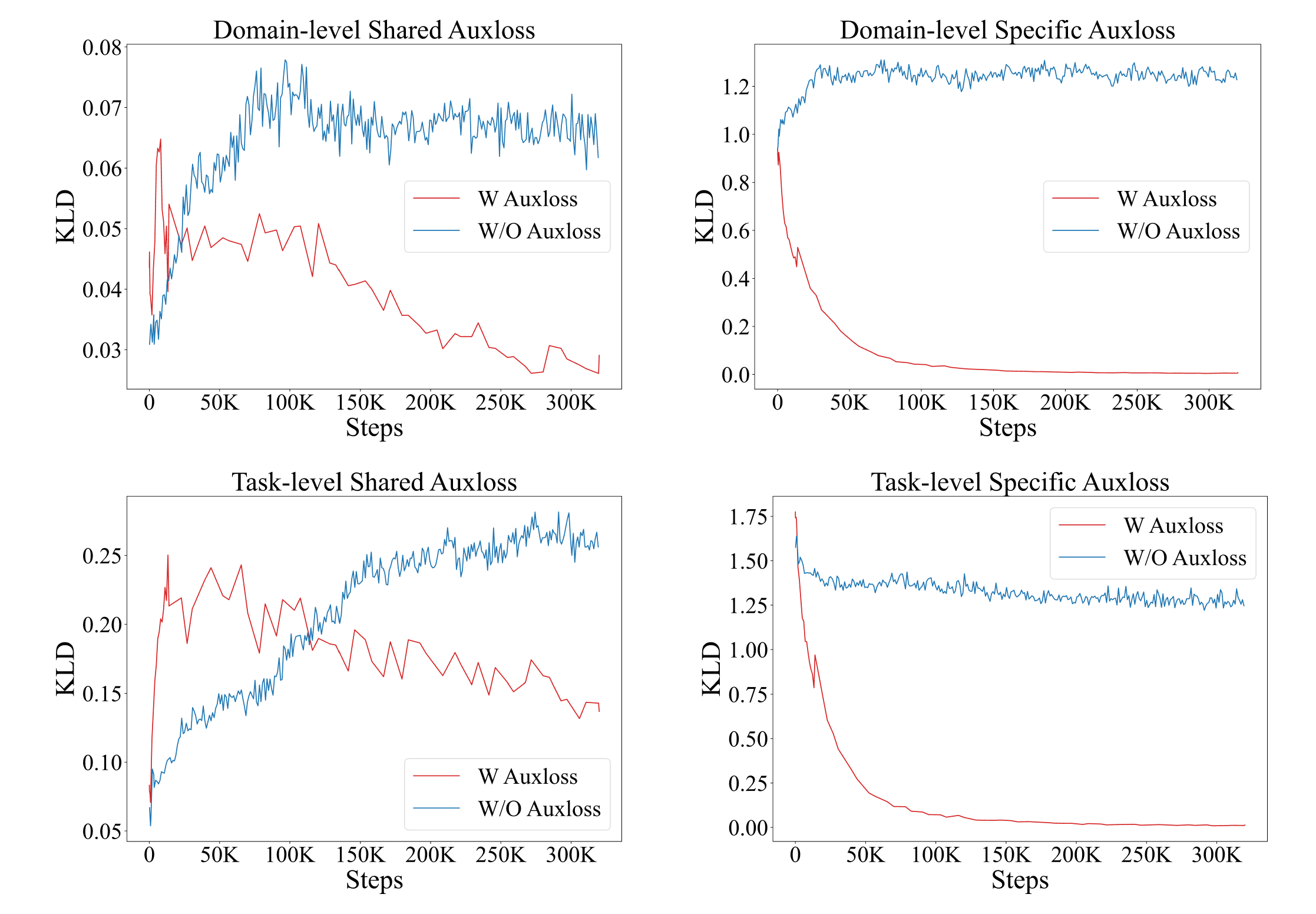}  
\caption{Kullback-Leibler divergence curves of both scenario and task levels on AliPay dataset. $X$-axis denotes training steps, $y$-axis is the summation of KL values between the assumed distribution and the selected row vector of gating matrix in Equation \ref{eqn:loss_sp}; lower $y$ value means closer distance to the assumed distribution.}  
\label{fig:kld}
\end{figure}

We conduct experiments to quantify scenario relationships in AliPay dataset. Specifically, we train a single Multi-task model (ESMM)~\cite{ESMM} for each scenario, and produce predictions for all scenarios during inference phase. The results of CTR and CTCVR on test set of each scenario are shown in Table~\ref{tab:SSL_ctr} and~\ref{tab:SSL_ctcvr} respectively.

It is intuitive that the best performance for a model is obtained by its own data. In other words, each scenario in AliPay dataset has its own pattern. Scenarios in a same domain type get closer than other scenarios by both CTR and CTCVR metrics. This result suggests that there exists inner similarities and differences among scenarios.

Compared with models trained in the multi-scenario\&multi-task setting (ref. Table~\ref{tab:res_alipay}), all baselines suffer from negative transfer from different scenarios. For instance, their performance on HP\&BS is inferior to the model trained on single scenario. However, our \baby outperforms all single-scenario models across all scenarios, which implies that our model can make better use of scenario relationships and avoid negative transfer.

\begin{table}[!]
  \caption{AUC results (CTCVR) of transfering to different scenarios.}
  \label{tab:SSL_ctcvr}
  \begin{tabular}{c|cccc}
    \toprule
    \diagbox{Train}{Test} &HP\&RI & HP\&BS & VP\&RI & VP\&BS \\
    \midrule
    \ HP\&RI & 91.18  & 65.36 &79.40  &65.51  \\
    \ HP\&BS & 53.25  & 82.11  &51.56  &65.64  \\
    \ VP\&RI & 87.35    & 53.62 &85.38  &66.13 \\
    \ VP\&BS &55.02      &73.26  &58.51  &74.04 \\
  \bottomrule
\end{tabular}
\end{table}

\begin{table}[!]
  \caption{The relative improvements on AliPay App with online A/B test. GMV refers to Gross Merchandise Volume.}
  \label{tab:abtest_all}
\begin{tabular}{ccccc}
\toprule
 \ Metric & CTR  & CVR & CTCVR & GMV\\
\hline
\ Relative Improved & +0.10\% & +2.61\% & +2.51\% & +7.21\% \\
\bottomrule
\end{tabular}
\end{table}

%=============================
\subsection{Ablation Study}
\label{ssec:expAblation}
%=============================

To study the impact of each component, we come up with two variants of \baby: (i) without Gaussian noise \& auxiliary loss, and (ii) without auxiliary loss, for ablation study. According to Table~\ref{tab:aba}, the simplest variant gets worst performance on all tasks for all scenarios. Gaussian noise $\bm{\eta}$ (ref. Equation~\ref{eqn:gj}) works well in all scenarios. This implies that a reasonable fluctuation would enable more freedom to explore  the expert selection space and achieve a more accurate structure.

As to the auxiliary loss $\mathcal{L}_{aux}$ (ref. Equation~\ref{eqn:loss_Laux}), it promotes the selecting procedure by guiding the gating scalar value of specific/shared experts network to get closer to the assumed specific/shared distribution. Table~\ref{tab:aba} shows that the performance across all the combinations of scenario and task degrade when removing $\mathcal{L}_{aux}$. We further plot the Kullback-Leibler divergence between the selected scenario/task distributions in the gating matrix and the assumed distribution (ref. Equation \ref{eqn:loss_sp}) in Figure~\ref{fig:kld}. It is obvious to see that \baby can produce sharper expert selection decisions by including $\mathcal{L}_{aux}$.

\begin{table}[!]
  \caption{The result of online A/B test in each single scenario.}
  \label{tab:abtest_single}
\begin{tabular}{c|cc}
\toprule
 \ Scenario & CTCVR & GMV\\
\hline
\ HP\&RI & +4.91\% & +12.97\% \\
\ HP\&BS & +5.84\% & +31.15\% \\
\ VP\&RI & +1.60\% & +4.86\% \\
\ VP\&BS & +8.78\% & +14.81\% \\
\bottomrule
\end{tabular}
\end{table}

\begin{figure}[!] 
\centering  
\includegraphics[width=0.48\textwidth]{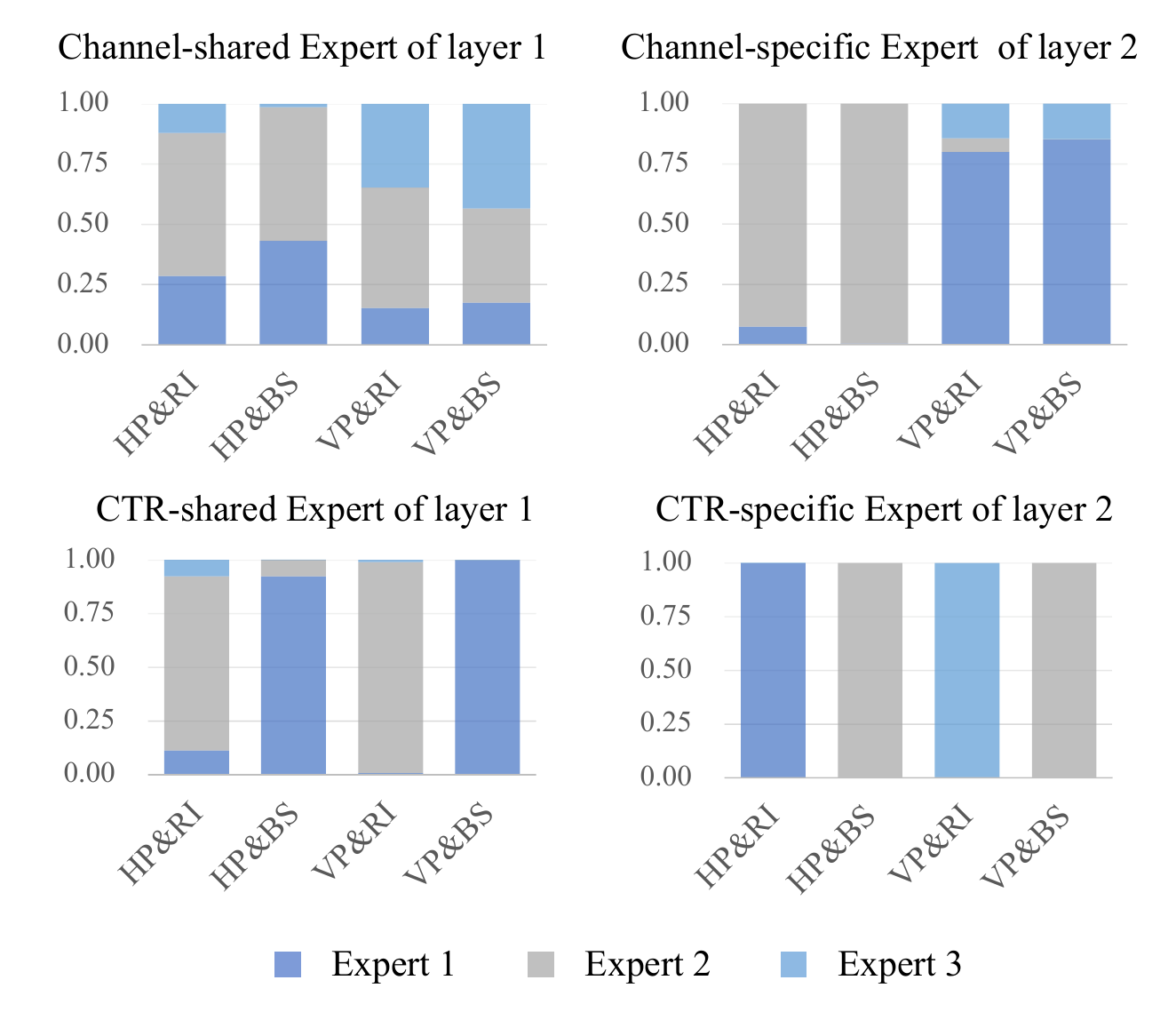}  
\caption{Expert utilization rate for being grouped into specific or shared group in different layer.} 
\label{fig:utili}
\end{figure}

%=============================
\subsection{Visualization of Expert Utilization}
\label{ssec:expVisual}
%=============================

We now investigate the expert utilization for both scenario/task-specific group and scenario/task-shared group respectively in AliPay dataset. For simplicity, every level (\ie channel, domain and task) contains two layers. Both specific expert set and shared expert set are configured to select only one expert. Figure~\ref{fig:utili} shows that in channel level, scenarios sharing the same type of channel have closer distributions. This indicates that our model can dynamically model complex patterns of shared information and disparity across scenarios. On the contrast, PLE defines specific/shared experts in a static way. At task level, we observe each task almost selects one certain experts in specific/shared group, meaning that our model can also converge to PLE setting. These observations suggest that \baby is a more general model which can adapt to different structure across scenarios/tasks.

%=============================
\subsection{Online A/B Test}
\label{ssec:expABTest}
%=============================

Currently, \baby is deployed on AliPay search system with the objective of maximizing CTCVR. We conduct a two-weeks'  online A/B test against the latest previously deployed search system on AliPay. Table~\ref{tab:abtest_all} shows the average relative improvements obtained by our model, suggesting \baby promotes online industrial applications on all metrics.

We further investigate the performance on single scenarios. From Table~\ref{tab:abtest_single}, we can observe that both CTCVR and GMV get considerable improvement on each single scenario. The different amounts of improvement are dependent on the scale of each scenario (ref. Table ~\ref{tab:alipay_stat}).

\section{CONCLUSION}
A large-scale search system consists of a series of fine-grained scenarios that share some inherent connections and bear distinct characteristics. The need to optimize diverse tasks across different scenarios in a cost-effective fashion is becoming increasingly urgent. In this paper, we propose a unified framework of integrating both MSL and MTL under an adaptive learning paradigm. Both offline and online evaluations well validate the superiority of our proposed \baby in terms of both search accuracy and business values. Also, \baby has been successfully deployed in the AliPay search platform for handling major traffic. We plan to extend the proposed \baby to support search explainability in the near future.

\begin{acks}
    This work was supported by National Natural Science Foundation of China (No.~61872278); and Young Top-notch Talent Cultivation Program of Hubei Province. Chenliang Li is the corresponding author.
\end{acks}

\bibliographystyle{ACM-Reference-Format}
\balance
\bibliography{refer}
\end{document}